\documentclass{article}
\usepackage{cite}
\usepackage{spconf,amsmath,epsfig}
\usepackage{xcolor}

\title{DisasterNets: Embedding Machine Learning in Disaster Mapping}
\name{Qingsong Xu\textsuperscript{1}, Yilei Shi\textsuperscript{2},  Xiao Xiang Zhu\textsuperscript{1}}
\address{1 Data Science in Earth Observation, Technical University of Munich (TUM), Munich, Germany\\
	2  Chair of Remote Sensing Technology (LMF), Technical University of Munich (TUM), Munich, Germany
}
\begin{document}
	%
	\maketitle
	\begin{abstract}
		Disaster mapping is a critical task that often requires on-site experts and is time-consuming. To address this,  a comprehensive framework is presented for fast and accurate recognition of disasters using machine learning, termed DisasterNets. It consists of two stages, space granulation and attribute granulation.  The space granulation stage leverages supervised/semi-supervised learning, unsupervised change detection, and domain adaptation with/without source data techniques to handle different disaster mapping scenarios. Furthermore, the disaster database with the corresponding geographic information field properties is built by using the attribute granulation stage.  The framework is applied to earthquake-triggered landslide mapping and large-scale flood mapping. The results demonstrate a competitive performance for high-precision, high-efficiency, and cross-scene recognition of disasters.  To bridge the gap between disaster mapping and machine learning communities, we will provide an openly accessible tool based on DisasterNets. The framework and tool will be available at \textcolor{blue}{https://github.com/HydroPML/DisasterNets}.
	\end{abstract}
	\section{Introduction}
	Disaster mapping is an essential task following tragic events such as hurricanes, earthquakes, and floods. It is also a time-consuming and risky task that still often requires the sending of experts on the ground to meticulously map and assess the damages. With a growing number of satellites in orbit (such as Sentinel, ASTER, and Landsat), it is easy to acquire almost real-time remote sensing images from areas struck by a disaster. However, it is challenging for real-time disaster mapping due to the massive amount of remote sensing data, variations in different disaster scenarios, and the time sensitivity for post-disaster rescue. To this end, leveraging deep learning, a comprehensive and general framework for disaster mapping, termed DisasterNets, is proposed for high precision and fast recognition of different disasters. 
	
	Specifically, the proposed framework includes five modules to achieve end-to-end disaster mapping under different scenarios.  When the corresponding labeled disaster training samples are available in the study area, a supervised/semi-supervised deep learning module is utilized for semantic segmentation in remote sensing images of disasters. For disaster mapping on synthetic aperture radar (SAR) or high-resolution remote sensing images, the main challenges of the supervised/semi-supervised deep learning module include high intra-class variance, low inter-class variance, the large variance of object scales, and dependency on training samples. To address these challenges, some supervised deep learning networks (such as MFFENet~\cite{xu2022mffenet,xu2020attention}) by fusing multi-scale features of objects and semi-supervised deep learning networks (such as SSCDNet~\cite{guo2022semi}) by the self-training technology are utilized in DisasterNets. In most cases, the suddenness of disasters and the massive amount of data in remote-sensing images make the labeling task difficult. Toward this end, an unsupervised change detection module using available pre-disaster remote sensing images, an unsupervised domain adaptation with source data module for existing source domain disaster datasets with labels, and an unsupervised domain adaptation without source data module for unavailable source domain disaster datasets are presented to segment the unlabeled post-disaster remote sensing images. The main difficulties affecting unsupervised change detection in remote sensing images are differences in light conditions, atmospheric conditions, and seasonality due to different acquisition dates. Thus, some unsupervised change detection networks (such as UCDFormer~\cite{xu2023UCDFormer}, DCVA~\cite{saha2019unsupervised}) by considering the distribution differences and computational costs are presented in DisasterNets. Furthermore, for the unsupervised domain adaptation with source data module, some deep learning networks (such as ADANet~\cite{xu2022mffenet}, CaGAN~\cite{9262039}) are used to address the domain adaptation problem in disaster mapping by leveraging the adversarial learning behaviors of GANs to perform distribution alignment in the pixel, feature, and output spaces of CNN networks. For the unsupervised domain adaptation without source data module, some source-free domain adaptation networks (such as SGD-MA~\cite{xu2023Uni}) are utilized in DisasterNets by generating a reliable synthetic source domain. Finally, the disaster database with the corresponding geographic information field properties obtained by using an attribute granulation module is presented for emergency rescue, risk assessment, and other applications. 
	
	The proposed DisasterNets is applied in earthquake-triggered landslide mapping and flood mapping under different scenarios. Specifically, high-resolution remote sensing images of three earthquake-induced landslides in Jiuzhaigou, Wenchuan (China) and Hokkaido (Japan), as well as  SAR remote sensing images of Pakistan flood are used to verify the practicality and effectiveness of DisasterNets under different scenarios. The results demonstrate a competitive performance for high-precision, high-efficiency, and cross-scene recognition of different disasters, and the insightful results are beneficial to disaster mapping based on machine learning methods.

	\begin{figure}[!htp]
		\centering
		{\includegraphics[width = 0.48\textwidth]{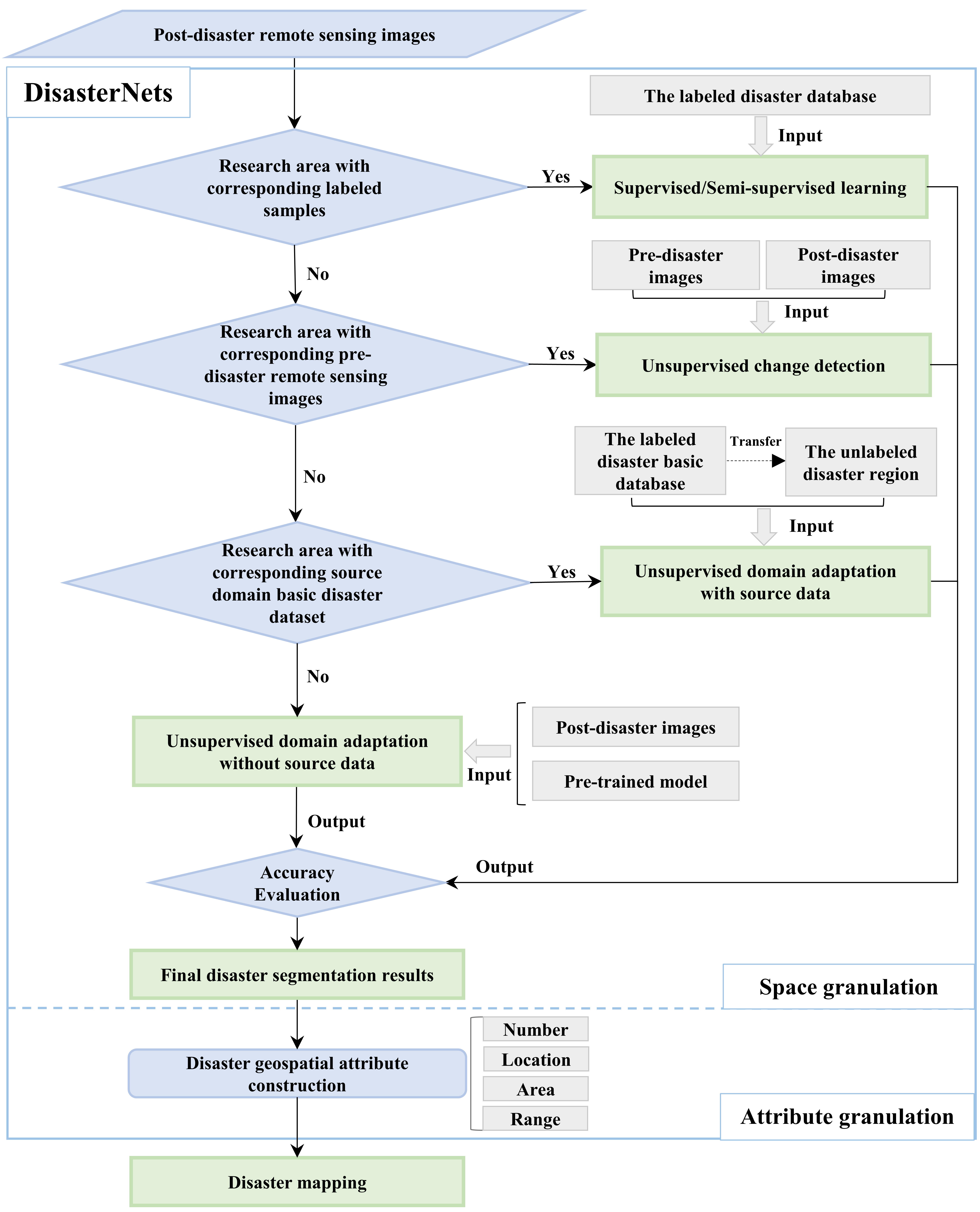}}
		\vspace{-2mm}
		\caption{Overview of the proposed DisasterNets.}
		\label{fig:0}
		\vspace{-4mm}
	\end{figure}
	
	\section{DisasterNets}
	As shown in Fig.~\ref{fig:0}, the DisasterNets consists of two stages, space granulation and attribute granulation. For the space granulation stage, the primary objective is to acquire disaster mapping. The attribute granulation stage focuses on incorporating geographical attributes to facilitate swift disaster assessments for each affected area.
	\vspace{-4mm}
	\subsection{Supervised/Semi-supervised Learning}
	\vspace{-2mm}
	When the corresponding labeled disaster training samples are available in the study area, a deep supervised segmentation
	model, dubbed Multiscale Feature Fusion with Encoder-decoder Network (MFFENet)~\cite{xu2022mffenet}, is utilized for large number of labeled samples, and a semi-supervised deep learning network (such as SSCDNet~\cite{guo2022semi}) is used for few labeled samples. Here, we will provide a brief introduction to MFFENet and SSCDNet.
	
	MFFENet consists of two parts: encoder and decoder. The encoder part is to generate different feature levels using the backbone network ResNet101. In addition, the atrous spatial pyramid pooling (ASPP) is used to generate context-reinforced features, and we modify the sampling rates of ASPP to $\{$6, 12, 24, 36$\}$ to produce denser feature maps with larger field-of-views. In the decoder part, an adaptive triangle fork (ATF) module is utilized to adaptively fuse the useful features at different scales and the same scales, and the
	residual convolution module is used as the basic processing unit. In addition, a dense top-down feature pyramid module is presented to gather more contextual information from the outputs of ATF and the encoder. Finally, to efficiently optimize the MFFENet model, a  boundary-aware loss  is introduced, which reweights the pixels near the boundary.
	Detailed descriptions of MFFENet, including its optimization,
	are presented in the work~\cite{xu2022mffenet}.

	SSCDNet incorporates the unsupervised domain adaptation strategy to address the domain shift problem in semi-supervised learning. By adapting at the feature level and output level, the network reduces domain distribution gaps and generates highly certain predictions for unlabeled samples. To improve the quality of pseudo-labels, a trustworthy pseudo-labeling method is employed, leveraging feedback information from output-level domain adaptation and a threshold strategy to identify trustworthy regions. These trustworthy regions are then used as ground-truth labels for network training. Further details of SSCDNet can be found in the work~\cite{guo2022semi}.
	\vspace{-4mm}
	\subsection{Unsupervised Change Detection}
	When there are no corresponding labeled samples in the research area, but there are pre-disaster remote sensing images, some unsupervised change detection networks (such as UCDFormer~\cite{xu2023UCDFormer}) are presented in DisasterNets.
	
	UCDFormer is a three-step approach for a scenario of unsupervised change detection with domain shift, which considers style differences and seasonal differences between
	multi-temporal images. Specifically, first, a
	transformer-driven image translation module is utilized to
	map data across two domains with real-time efficiency. A light-weight transformer is
	proposed in the transformer-driven image translation module
	to reduce the computational complexity of the self-attention
	layer in regular transformers by using spatial
	downsampling and channel group operations. In addition, a priori change indicator,
	affinity weight, is intended to reduce the translation
	strength of changed pixels and to increase the translation
	of unchanged pixels, by using a weighted translation loss. Next, a reliable
	pixel extraction module is proposed to extract significantly
	changed/unchanged pixel positions from the difference map,
	which is computed by fusing multi-scale feature maps. Finally, a binary change map of disasters is obtained based on the reliable changed/unchanged pixel positions and the random forest
	classifier. Detailed descriptions of UCDFormer are provided in  the work~\cite{xu2023UCDFormer}.
	\vspace{-4mm}
	\subsection{Unsupervised Domain Adaptation}
	\vspace{-2mm}
	In most cases, the suddenness of disasters and the massive
	amount of data in remote sensing images make the labeling task difficult. Toward this end, 
	some unsupervised domain adaptation (UCD) networks (such as an adversarial
	domain adaptation network (ADANet)~\cite{xu2022mffenet}, a class-aware generative
	adversarial network (CaGAN)~\cite{9262039}) are used to address the domain adaptation with source data module problem in disaster mapping by leveraging the adversarial learning behaviors of GANs to perform distribution alignment in the pixel, feature, and output spaces of CNN networks. In addition, some source-free domain adaptation networks (such as SGD-MA~\cite{xu2023Uni}) are utilized in DisasterNets for the domain adaptation without source data problem, by generating a reliable synthetic source domain.
	
	For a learnable and life-long model to perform different disaster segmentation tasks, it should be able to reutilize the information acquired in previous disaster segmentation tasks
	with the labeled images and transfer it to the new learning tasks of
	disaster segmentation with no labeled images. ADANet is proposed to mitigate
	the domain shift in different data distributions of disasters. ADANet consists
	of two modules: a segmentation network and two discriminators.
	Some supervised encoder-decoder architectures (such as MFFENet) can be utilized as the generator. Hence, the output features
	of the encoder block and the decoder block in the
	network are both collected and adopted since the former contains
	rich overall semantic information and the latter contains rich context, scene layout, and other detailed information. Detailed descriptions of ADANet are provided in  the work~\cite{xu2022mffenet}.
	
	However, the source data is usually not accessible in many cases due to privacy or disaster emergency urgency. Thus, UDA without source data (SDG-MA) is utilized in DisasterNets. SDG-MA consists of a source
	data generation (SDG) stage and a model adaptation (MA)
	stage. In the SDG stage, we reformulate the goal as estimating
	the conditional distribution rather than the distribution of the
	source data, since the source data space is exponential with
	the dimensionality of data. After the conditional distribution
	of the source data is obtained, it becomes a UDA task
	to mitigate the domain shift of different disaster data. In addition, a novel transferable weight in SDG-MA is defined by
	considering confidence and domain similarity to distinguish different categories in each domain. For more details about SDG-MA, please refer to the work~\cite{xu2023Uni}.
	\vspace{-4mm}
	\subsection{Attribute Granulation}
	Following the spatial granulation of the regional disasters, the next step involves calculating the area, perimeter, aspect ratio, and centroid of each disaster. Additionally, spatial identification is performed on the disasters detected in the remote sensing imagery, and the corresponding geographical coordinates of their centroids (locations) are incorporated. Utilizing a geographic information system, these characteristics of the disasters are defined as attribute properties, thereby completing the construction of their spatial geographic information attributes. Finally, the regional disaster database is rapidly established.
	\vspace{-4mm}
	\section{Experiments}
	\vspace{-2mm}
	\begin{figure}[!htp]
		\centering
		{\includegraphics[width = 0.45\textwidth]{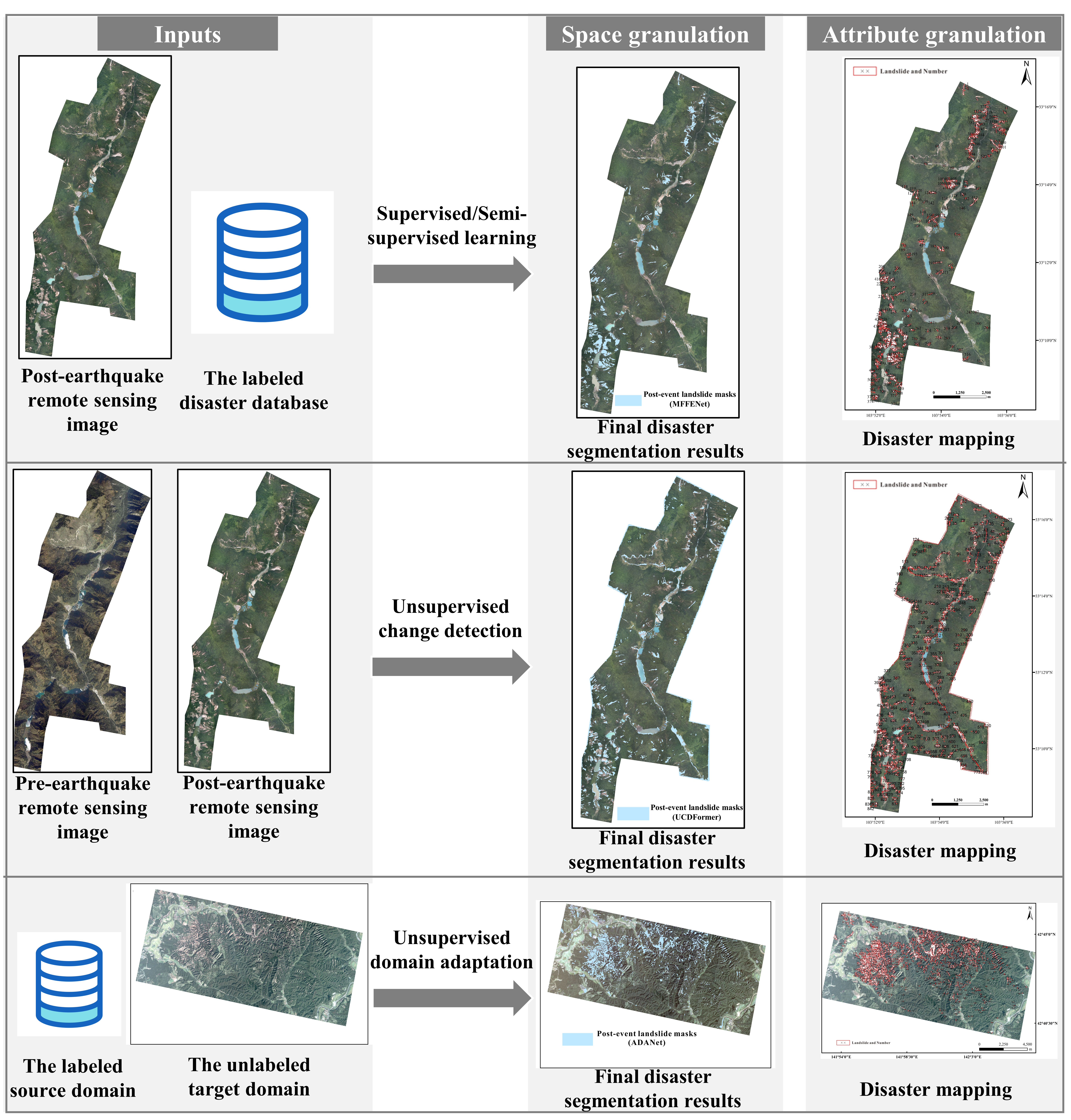}}
		\vspace{-2mm}
		\caption{Earthquake-triggered landslide mapping using DisasterNets.}
		\label{fig:3}
		\vspace{-4mm}
	\end{figure}
	
	\textbf{Datasets.} To assess the performance of the scheme for different disasters, two experiments are tested, including earthquake-triggered landslide mapping and   large-scale flood mapping. 
	
	For earthquake-triggered landslide mapping using RGB images, first, when the labeled landslide database~\cite{xu2022mffenet} is accessible, the supervised method (MFFENet) is applied to the earthquake-induced Jiuzhaigou landslides.  The research region chosen in the Jiuzhaigou earthquake covers an area of 53.6 km$^{2}$. Then, with the availability of pre-earthquake remote sensing images of Jiuzhaigou, the unsupervised change detection module, such as UCDFormer, is employed to identify the earthquake-induced landslides in the region. Next, the earthquake-induced landslides in Hokkaido, Japan, are chosen as a case study to assess the effectiveness of UCD with source data, using the labeled landslide database as the source data. Notably, a representative small region of the earthquake-induced Hokkaido landslides is carefully selected to evaluate the agreement between the ground truth obtained through visual interpretation and the predicted results generated by the ADANet model. Finally, the earthquake-induced landslides in Wenchuan, China are selected as a case study to assess the effectiveness of UCD without source data.  Furthermore,   the Pakistan flood in 2022 (30.49 km$^{2}$) is employed to verify the effectiveness of DisasterNets. Given that the flood in Pakistan is classified as an open area flood event, we directly utilize SAR data from the Sentinel-1 satellite to acquire pre-change and post-change SAR images of floods. Then, UCDFormer is used to swiftly generate a map of the affected flood area.
	\begin{table}[!htp]
		\vspace{-3mm}
		\caption{Results of earthquake-triggered landslide mapping using DisasterNets.}
		\centering
		\resizebox{0.5\textwidth}{!}
		{
			\centering
			\begin{tabular}{cccccc}
				\hline
				Methods                                     & Datasets                                                 & Precision & Recall         & OA    & Time (Training+Inference) \\ \hline
				Supervised Learning              & Jiuzhaigou (53.6 km$^{2}$) & 90.23     & 92.44          & 99.70 & about 12 hours            \\ \hline
				Unsupervised Change Detection & Jiuzhaigou (53.6 km$^{2}$) & 48.04     & 25.00 & 96.22 & about 10 minutes          \\ \hline
				Unsupervised Domain Adaptation   & Hokkaido (159.3 km$^{2}$)  & 89.72     & 87.06          & 89.43 & about 7.3 hours              \\ \hline
		\end{tabular}}
	\vspace{-4mm}
		\label{Tab1}
	\end{table}
	\begin{figure}[!htp]
		\centering
		{\includegraphics[width = 0.4\textwidth]{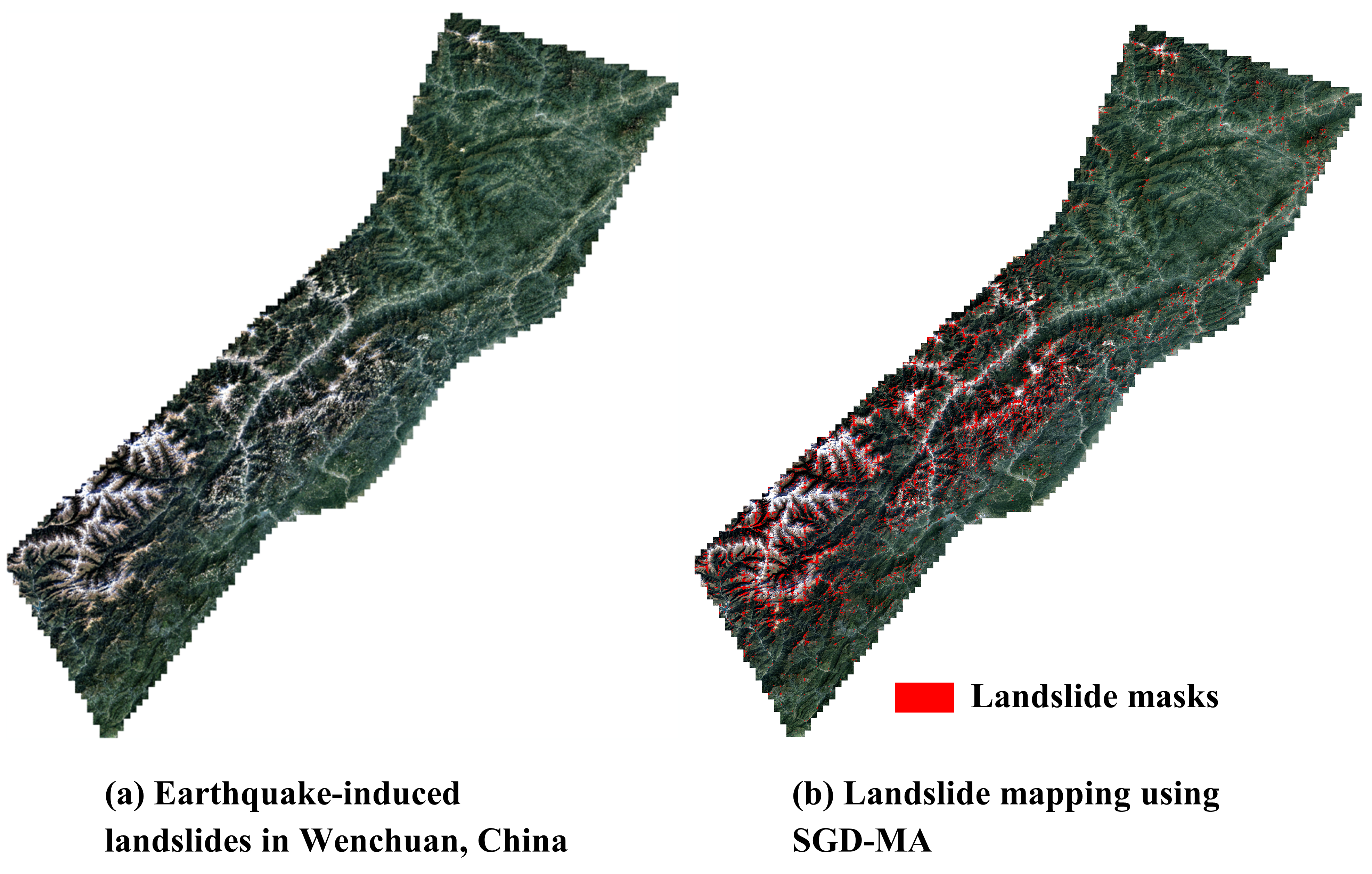}}
		\vspace{-2mm}
		\caption{Earthquake-triggered Wenchuan landslide mapping using UCD without source data.}
		\label{fig:4}
		\vspace{-4mm}
	\end{figure}

	\textbf{Results of earthquake-triggered landslide mapping.} According to the visualized results in Fig.~\ref{fig:3} and quantitative results in Table~\ref{Tab1}, it is evident that the supervised method yields the highest accuracy, followed by UDA with source data, and finally unsupervised change detection.  However, the situation is reversed when considering training and inference time. Unsupervised change detection achieves a precision of 48.04\% within approximately 10 minutes, whereas the supervised method achieves a precision of 90.23\% but requires around 12 hours for processing. Additionally, the visualization results in Fig.~\ref{fig:4} indicate that SGD-MA can achieve satisfactory recognition even without the use of source data.
	\begin{figure}[!htp]
		\centering
		{\includegraphics[width = 0.45\textwidth]{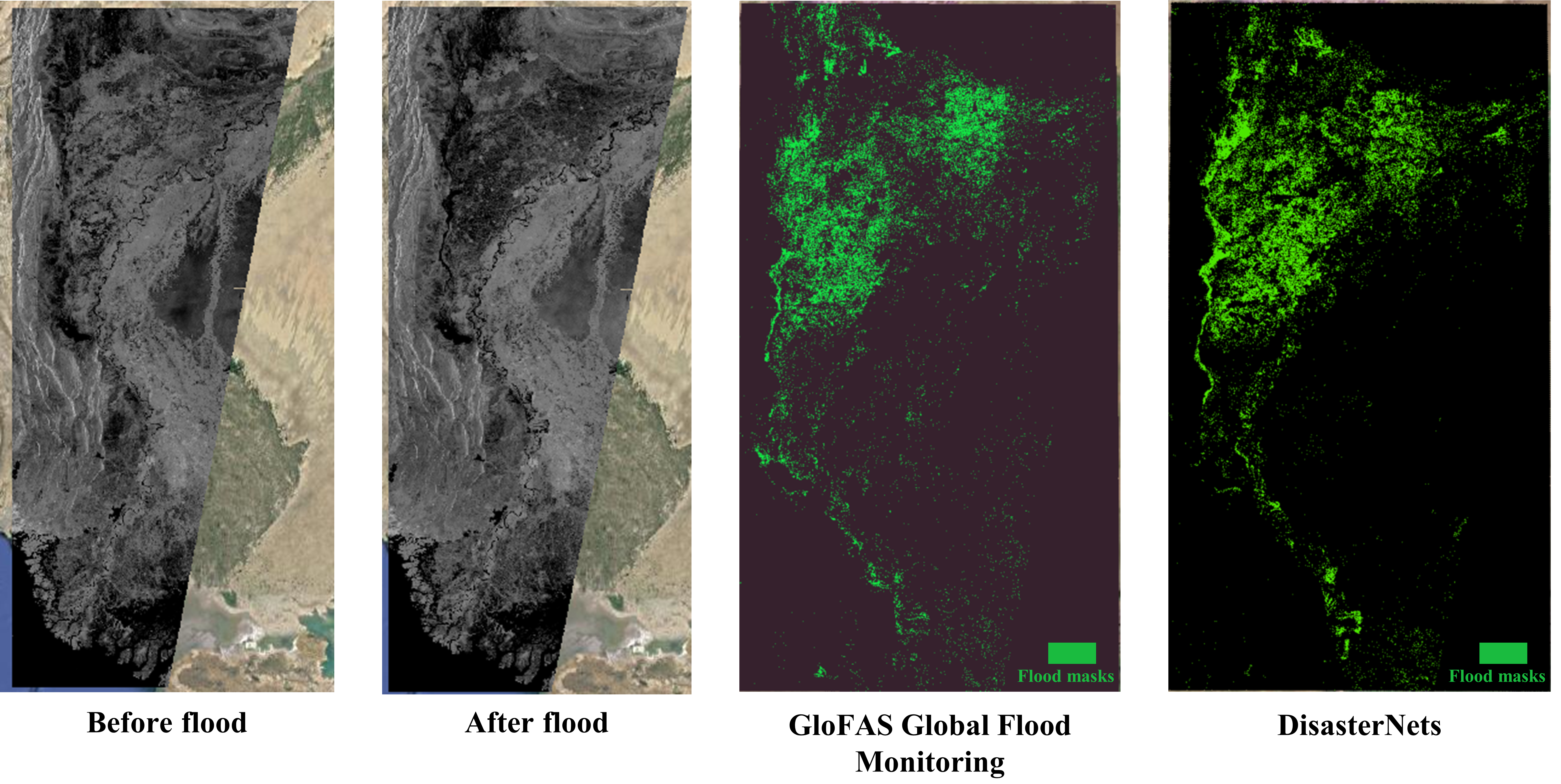}}
		\vspace{-2mm}
		\caption{Results of Pakistan flood mapping using unsupervised change detection (UCDFormer).}
		\label{fig:5}
		\vspace{-4mm}
	\end{figure}
	
	\textbf{Results of Pakistan flood mapping.} According to the visualized flood mapping of Pakistan 2022 in Fig.~\ref{fig:5}, our results using unsupervised change detection (UCDFormer) exhibit a high degree of consistency with the  public results from GloFAS global flood monitoring. This strong agreement highlights the effectiveness of our DisasterNets in rapidly mapping flood hazards.
	\vspace{-4mm}
	\section{Conclusion}
	\vspace{-4mm}
	In this study,  a comprehensive framework for fast and accurate disaster recognition using machine learning, DisasterNets, is proposed. The framework, consisting of space granulation and attribute granulation stages, demonstrates competitive performance in earthquake-triggered landslide mapping and large-scale flood mapping. 
	\vspace{-4mm}
	\bibliographystyle{IEEEbib}
	\bibliography{strings,refs}

\begin{thebibliography}{1}

\bibitem{xu2022mffenet}
Qingsong Xu, Chaojun Ouyang, Tianhai Jiang, Xin Yuan, Xuanmei Fan, and Duoxiang
  Cheng,
\newblock ``Mffenet and adanet: a robust deep transfer learning method and its
  application in high precision and fast cross-scene recognition of
  earthquake-induced landslides,''
\newblock {\em Landslides}, pp. 1--31, 2022.

\bibitem{xu2020attention}
Qingsong Xu, Xin Yuan, Chaojun Ouyang, and Yue Zeng,
\newblock ``Attention-based pyramid network for segmentation and classification
  of high-resolution and hyperspectral remote sensing images,''
\newblock {\em Remote Sensing}, vol. 12, no. 21, pp. 3501, 2020.

\bibitem{guo2022semi}
Jianhua Guo, Qingsong Xu, Yue Zeng, Zhiheng Liu, and Xiaoxiang Zhu,
\newblock ``Semi-supervised cloud detection in satellite images by considering
  the domain shift problem,''
\newblock {\em Remote Sensing}, vol. 14, no. 11, pp. 2641, 2022.

\bibitem{xu2023UCDFormer}
Qingsong Xu, Yilei Shi, Jianhua Guo, Chaojun Ouyang, and Xiaoxiang Zhu,
\newblock ``Ucdformer: Unsupervised change detection using real-time
  transformers,''
\newblock 2023.

\bibitem{saha2019unsupervised}
Sudipan Saha, Francesca Bovolo, and Lorenzo Bruzzone,
\newblock ``Unsupervised deep change vector analysis for multiple-change
  detection in vhr images,''
\newblock {\em IEEE Transactions on Geoscience and Remote Sensing}, vol. 57,
  no. 6, pp. 3677--3693, 2019.

\bibitem{9262039}
Qingsong Xu, Xin Yuan, and Chaojun Ouyang,
\newblock ``Class-aware domain adaptation for semantic segmentation of remote
  sensing images,''
\newblock {\em IEEE Transactions on Geoscience and Remote Sensing}, vol. 60,
  pp. 1--17, 2020.

\bibitem{xu2023Uni}
Qingsong Xu, Yilei Shi, Xin Yuan, and Xiaoxiang Zhu,
\newblock ``Universal domain adaptation for remote sensing image scene
  classification,''
\newblock {\em IEEE Transactions on Geoscience and Remote Sensing}, 2023.

\end{thebibliography}
	
\end{document}